\documentclass{article}
\usepackage{spconf,amsmath,epsfig}
\usepackage{tabularx}
\usepackage{subfig}
\usepackage{booktabs}
\usepackage{tabularx}
\newcommand{\etal}{\textit{et al.}}
\title{Semi-Supervised Object Detection for Sorghum Panicles in UAV Imagery}
\name{Enyu Cai, Jiaqi Guo, Changye Yang, and Edward J. Delp}
\address{Video and Image Processing Laboratory (VIPER)\\
School of Electrical and Computer Engineering\\
Purdue University\\
West Lafayette, Indiana, USA}
\begin{document}
%\ninept
%
\maketitle
\begin{abstract}
The sorghum panicle is an important trait related to grain yield and plant development.
Detecting and counting sorghum panicles can provide significant information for plant phenotyping.
Current deep-learning-based object detection methods for panicles require a large amount of training data.
The data labeling is time-consuming and not feasible for real application.
In this paper, we present an approach to reduce the amount of training data for sorghum panicle detection via semi-supervised learning.
Results show we can achieve similar performance as supervised methods for sorghum panicle detection by only using 10\% of original training data.
\end{abstract}
\section{Introduction}
Plant phenotyping is used to find the connection between a plant's physical properties and the genetic information \cite{walter_2015}.
Modern high-throughput plant phenotyping uses Unmanned Aerial Vehicles (UAVs) equipped with multiple sensors to collect imagery \cite{chapman_2014}.
The images collected by UAVs can be analyzed to estimate plant traits \cite{habib_improving_2016}.
Sorghum (\textit{Sorghum bi-color} (L.) Moench) is an important crop for food and biofuel production \cite{sweet_sorghum}.
The Sorghum panicle is a cluster of grains at the top of the plants that is critical to plant growth and management \cite{sweet_sorghum}.
Detecting sorghum panicles can help plant breeders estimate plant properties such as flowering time \cite{cai_2021}.
The deep neural network has shown successful results in general object detection tasks \cite{fasterrcnn}.
Recently, deep neural networks have also demonstrated the capability for detection tasks related to plant phenotyping \cite{lin_2020}.
However, a large amount images are needed for training the neural network.
Ground truthing a large amount of RGB images captured by UAVs is a major bottleneck relative to the performance of the detection tasks.
Semi-supervised classification approaches train the network with a small amount of labeled data and a large amount of unlabeled data to reduce the manual data labeling \cite{semi}.
The use of pseudo-labels \cite{lee_2013} is the key idea for semi-supervised approaches.
The pseudo-labels are the data labels generated by the model pretrained on the small dataset.
The pseudo-labels are combined with the real labels to expand the training dataset.
A semi-supervised loss is also introduced for training on labeled and unlabeled data.
Recent work focuses on regulating the loss function to maintain consistency during training.
MixMatch \cite{mixmatch} is an example of consistency regulation.
It uses data augmentation, label guessing, and MixUp on both labeled and unlabeled images.
FixMatch \cite{fixmatch} is another consistency-based method for semi-supervised classification.
It combines consistency regularization and pseudo-labeling to improve the performance of semi-supervised learning.
Similar to semi-supervised classification, pseudo-label-based approaches are used for semi-supervised object detection \cite{sohn_2020,soft_teacher, liu_2021}.
The approach consists of a teacher model and a student model.
The teacher model is trained with a small amount of data at first.
The teacher network will then generate the annotation from the unlabeled dataset to produce pseudo-labels.
The pseudo-labeled data and labeled data are combined to train the student model.
In \cite{sohn_2020}, Sohn \etal~introduce a framework, STAC, to generate highly confident pseudo labels and update the models by enforcing consistency.
STAC generates pseudo-labels from unlabeled data using non-maximum suppression (NMS).
The confidence-based method is used to filter pseudo-labels.
Unbiased Teacher \cite{liu_2021} is another framework to jointly train the student and teacher networks in a mutually-beneficial manner.
In this paper, we present a method to train a sorghum panicle detection deep neural network on RGB UAV images with a small amount of training data using semi-supervised learning.

\section{Methods}
\begin{figure}[htb]
	\centering
	\includegraphics[width=0.45\textwidth]{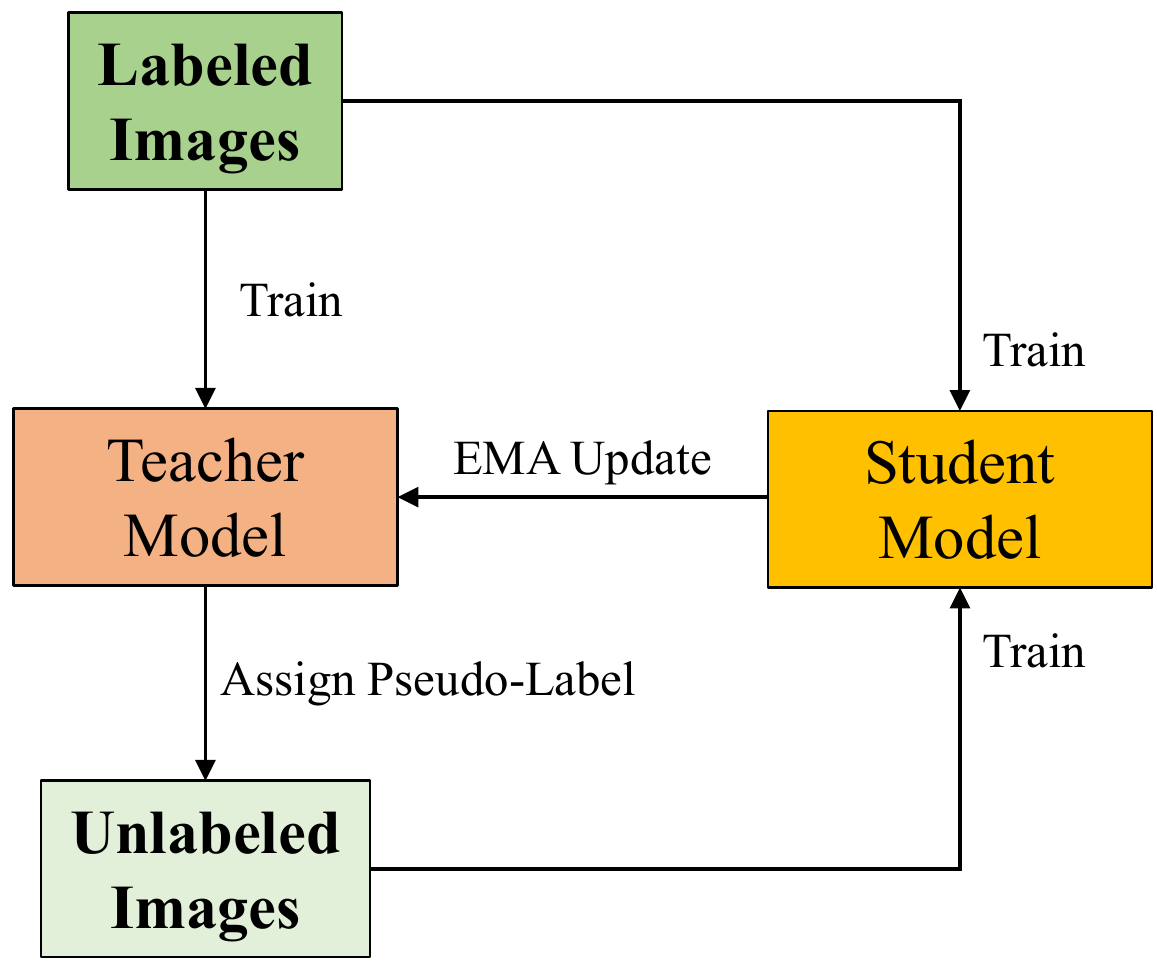}
	\caption{Block diagram of our semi-supervised learning framework.}
	\label{fig:ssod}
\end{figure}

We investigate  semi-supervised learning  for two-stage and one-stage 
object detection methods.
For two-stage object detection, we use the Soft Teacher \cite{soft_teacher} framework with Faster-RCNN.
For one-stage object detection, we choose the Efficient Teacher \cite{xu2023efficient} framework with YOLOv5.
The selection of the detection network is based on the performance of general detection datasets such as COCO \cite{coco_dataset}.
Theoretically, both semi-supervised methods are interchangeable with the other object detection method.
However, the performance is degraded if we simply apply one method to another due to the structure difference between one-stage networks and two-stage networks.
In this case, we choose semi-supervised methods that have the best fit for each type of detection network as a fair comparison.
The block diagram of our semi-supervised framework is shown in Figure \ref{fig:ssod}.

\subsection{Soft Teacher Framework}
The Soft Teacher framework consists of a teacher model and a student model.
The teacher model is trained using a small batch of labeled data and performs pseudo-labeling on the unlabeled images.
The student model is trained on both labeled and pseudo-labeled images.
During the training process, the teacher model is continuously updated by the student model through an exponential moving average (EMA) strategy.
The loss function of the Soft Teacher is a combined loss function from the supervised and unsupervised loss:
\begin{equation}
    \mathcal{L}=\mathcal{L}_{s}+\alpha \mathcal{L}_{u}
\end{equation}

\begin{equation}
    \mathcal{L}_{s}=\frac{1}{N_{l}}(\mathcal{L}_{cls}(I_{l}^{i})+\mathcal{L}_{reg}(I_{l}^{i}))
\end{equation}

\begin{equation}
    \mathcal{L}_{u}=\frac{1}{N_{u}}(\mathcal{L}_{cls}(I_{u}^{i})+\mathcal{L}_{reg}(I_{u}^{i}))
\end{equation}
where $\mathcal{L}$ is the weighted sum of supervised loss $\mathcal{L}_{s}$ and unsupervised loss $\mathcal{L}_{s}$, $\alpha$ is the weight for unsupervised loss, $\mathcal{L}_{cls}$ is the classification loss, $\mathcal{L}_{reg}$ is box classification loss, $I_{u}^{i}$ is the $i$-th unlabeled image, $I_{l}^{i}$ is the $i$-th labeled image, $N_{u}$ is the number of unlabeled image and $N_{l}$ is the number of labeled image.
During pseudo-label generation, the NMS and FixMatch \cite{fixmatch} strategy is used to remove duplicate bounding box candidates.
The high threshold value is also used for pseudo-label generation to improve the quality of pseudo-labels.
The process of pseudo-label generation will introduce error since some foreground box candidates will be assigned as negative.
To compensate for this problem, the Soft Teacher  introduces a loss function that uses more information from the teacher model.
The Soft Teacher framework also uses a jittering box refinement technique to filter out duplicate boxes.
The original method from Soft Teacher is training the teacher model and student model at the same time with random weights at the beginning.
In practice, we found the training is not stable due to the limited amount of images in our dataset.
We introduce another warm-up stage for the teacher model.
During the warm-up stage, the teacher model will be trained only with labeled data.
The trained weight will then be loaded into the co-training stage with the student model.

\subsection{Efficient Teacher Framework}
One-stage object detection networks \cite{yolov5} generally have higher recall and faster training speed compared to two-stage object detection networks \cite{fasterrcnn}.
However, the semi-supervised learning approach from two-stage detection networks is facing challenges when directly applied to a one-stage detection network.
The multi-anchor strategy used in the one-stage network magnifies the label imbalance problem from semi-supervised learning in the two-stage network, resulting in low-quality pseudo-labels and poor training results.
Efficient Teacher is a semi-supervised learning approach optimized for single-stage object detection networks.
To leverage the label inconsistency problem, Efficient Teacher introduces a novel pseudo-label assigner to prevent interference from low-quality pseudo-labels.
During training, each pseudo-label is assigned a pseudo-label score that represents the uncertainty of the label.
Two threshold value of the score $\tau_{1}$ and $\tau_{2}$ is used.
If a pseudo-label has a score between $\tau_{1}$ and $\tau_{2}$, the pseudo-label is categorized as an uncertain label.
The loss of uncertain labels is filtered out to improve the performance.
The Efficient Teacher framework also introduces epoch adaptor mechanism to stabilize and accelerate the training process.
Epoch adaptor combines domain adaptation and distribution adaptation techniques.
Domain adaptation enables training on both unlabeled data and labeled data during the first Burn-In phase to prevent overfitting on labeled data.
The distribution adaptation technique dynamically updates the thresholds $\tau_{1}$ and $\tau_{2}$ at each epoch to reduce overfitting.

\section{Experimental Results}
\begin{figure}[htbp]
  \centering\includegraphics[width=0.4\linewidth]
  {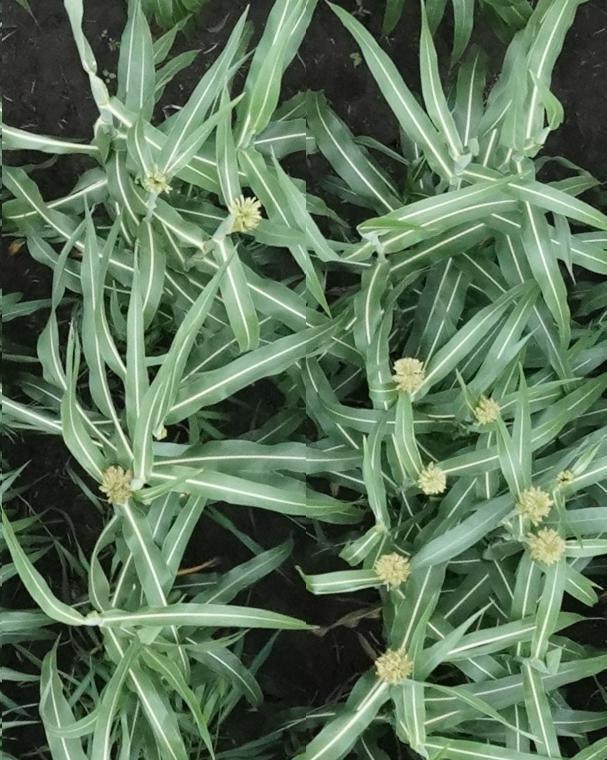}
  \centering\includegraphics[width=0.4\linewidth]{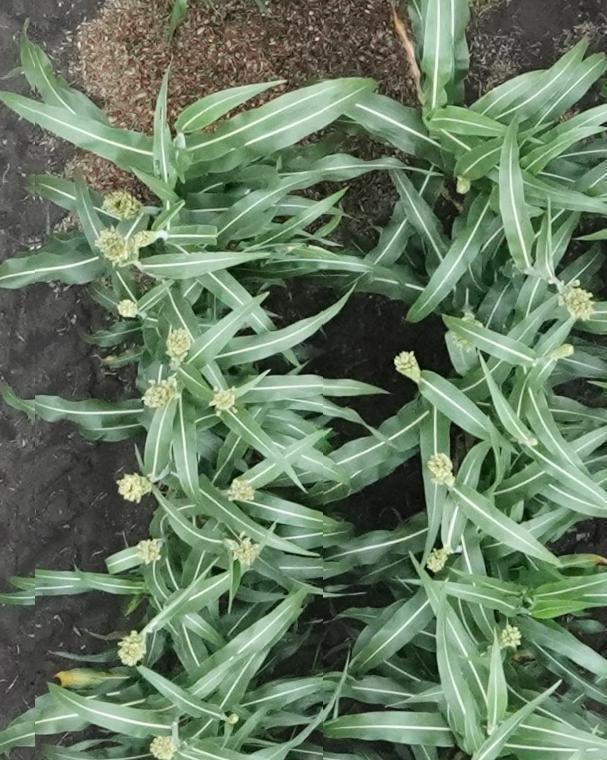}
  \caption{Sample images from the training dataset.}
  \label{fig:semi_data}
\end{figure}

The sorghum panicle dataset is from an RGB orthomosaic \cite{habib_improving_2016} captured by UAVs in a sorghum field.
We select a small region of the orthomosaic and crop it into small images for data labeling and training purposes as shown in Figure \ref{fig:semi_data}.
We select the early sorghum growing stage for the experiments.
Compared to the later stage, the early-stage sorghum panicles have more variation in shapes and sizes which brings more challenge to the methods.
Moreover, the fewer panicles in each image can further reduce the number of available labels for training.
In total, we have 364 images for training, 90 images for validation, and 60 images for testing.
Each image is resized to 640 $\times$ 640 resolution during training.
These RGB images are used to form a supervised baseline to compare with semi-supervised learning. 
For semi-supervised learning, we randomly select 1\%, 5\%, and 10\% within the training dataset to form a semi-supervised learning dataset.
The labels of the rest of the training data are removed correspondingly to represent the unlabeled data.
We have 3 labeled images for the 1\% dataset, 18 labeled images for the 5\% dataset, and 36 labeled images for the 10\% dataset.
Setting the appropriate training parameters is very important for semi-supervised learning on the limited dataset.
From the empirical experiment, we found the learning rate and NMS threshold for pseudo-labels have the most impact on training performance.
In the supervised learning stage, the learning rate can have a greater learning rate for fast convergence.
In the semi-supervised learning stage, the learning rate needed to be decreased for a very small amount of labeled data.
In practice, we found setting the learning rate of 0.001 for the supervised stage is appropriate for the warm-up of the teacher model.
The default semi-supervised learning rate from both methods is too large, resulting in unstable training.
We found to set the learning rate to 0.00005 is suitable for the semi-supervised stage in both methods.
For the NMS threshold in Efficient Teacher, we set the confidence threshold to 0.5 to reduce the false positive and the IoU threshold to 0.1 to reduce the duplicated bounding box.

We evaluate the semi-supervised learning approach by using three different settings of the original training dataset: 1\%, 5\%, and 10\% training data.
In the warm-up stage, we first trained the network with only 1\%, 5\%, and 10\% labeled data in a supervised manner to form a baseline.
In the semi-supervised stage, the weights of the baseline model are loaded into the teacher model.
The teacher model with pre-loaded weight is trained with the student model together.
For the Soft Teacher framework, we use Faster-RCNN \cite{fasterrcnn} with ResNet-50 \cite{resnet} backbone.
For the Efficient Teacher framework, we use the YOLOv5l \cite{yolov5} model.
The result of the Soft Teacher method is shown in Table \ref{tab:soft}, the semi-supervised learning increases the mAP by 5.6\% in 1\% labeled data, 2.5\% in 5\% labeled data, and 3.7\% in 10\% labeled data.
The result of the Efficient Teacher method is shown in Table \ref{tab:efficient}, the semi-supervised learning increases the mAP by 3.1\% in 1\% labeled data, 1.7\% in 5\% labeled data and 1.7\% in 10\% labeled data.
The Efficient Teacher method achieves the highest mAP due to the better YOLOv5 model in the baseline.
However, the Soft Teacher framework has the highest mAP increases in three scenarios.
The training is done on a single NVIDIA RTX A40 GPU.
The Soft Teacher took 7 hours to finish training while the Efficient Teacher only took one hour to finish.
Compare to supervised learning using fully labeled data (364 images), we can achieve comparable results with only 10\% of the original amount (36 images).

\begin{table}[t]
\centering
\begin{tabular}[c]{ccc}
\toprule
\textbf{mAP@[.5:.95]}  &  \textbf{Baseline}  &  \textbf{Soft Teacher} \\ \midrule
1\% & 38.2 &  \textbf{43.8}  \\
5\% & 42.8 & \textbf{45.3} \\
10\% & 43.4 & \textbf{47.1} \\
100\% & 50.2 &  \\
\bottomrule
\end{tabular}
\caption{Results from Soft Teacher Framework with Faster-RCNN. The baseline is supervised learning only with 1\%, 5\%, and 10\% data accordingly.}
\label{tab:soft}
\end{table}

\begin{table}[t]
\centering
\begin{tabular}[c]{ccc}
\toprule
\textbf{mAP@[.5:.95]}  &  \textbf{Baseline}  & \textbf{Efficient Teacher} \\ \midrule
1\% & 38.1 &  \textbf{41.2}  \\
5\% & 45.9 & \textbf{47.6} \\
10\% & 47.4 & \textbf{49.1} \\
100\% & 51.2 &  \\
\bottomrule
\end{tabular}
\caption{Results from Efficient Teacher Framework with YOLOv5. The baseline is supervised learning only with 1\%, 5\%, and 10\% data accordingly.}
\label{tab:efficient}
\end{table}

\section{Conclusion and Discussion}

In this paper, we propose a method for reducing training data in sorghum panicle detection.
We examine two different types of semi-supervised learning approaches for sorghum panicle detection.
We demonstrate the capability of semi-supervised learning methods for achieving similar performance by only using 10\% of training data compared to the supervised approach.
Future work includes developing auto-tuning methods for the hyper-parameters and extending the methods to other plant traits.

\section{Acknowledgments}
We thank Professor Ayman Habib and the Digital Photogrammetry Research Group (DPRG) from the School of Civil Engineering at Purdue University for providing the images used in this paper. The work presented herein was funded in part by the Advanced Research Projects Agency-Energy (ARPA-E), U.S. Department of Energy, under Award Number DE-AR0001135.
The views and opinions of the authors expressed herein do not necessarily state or reflect those of the United States Government or any agency thereof.
Address all correspondence to Edward J. Delp, ace@ecn.purdue.edu

% References should be produced using the bibtex program from suitable
% BiBTeX files (here: strings, refs, manuals). The IEEEbib.bst bibliography
% style file from IEEE produces unsorted bibliography list.
% -------------------------------------------------------------------------
\bibliographystyle{IEEEbib}
\bibliography{refs}

\end{document}